%% file: root.tex
\documentclass[letterpaper, 10 pt, conference]{Packages/ieeeconf}  
\usepackage{microtype}  
\microtypesetup{protrusion=true,expansion=true}

\IEEEoverridecommandlockouts                              
 
\usepackage{float}
\usepackage{times}
\usepackage{epsfig}
\usepackage{graphicx}
\usepackage{amsmath}
\usepackage{amssymb}
\usepackage[ruled]{algorithm}
\usepackage{algpseudocode}
\usepackage{ctable}
\usepackage[export]{adjustbox}
\usepackage{import}
\usepackage{pgf}
\usepackage{setspace}
\usepackage{url}
\usepackage{hyperref}
\usepackage{capt-of,etoolbox}
\usepackage{pifont} 
\usepackage{mathtools}
\newcommand{\xmark}{\ding{55}}%

\usepackage{fontawesome} 
\usepackage{multicol}  
\usepackage{balance}
\usepackage{cuted}

\makeatletter
\let\NAT@parse\undefined
\makeatother
\usepackage[sort&compress,numbers]{natbib}

\usepackage{Packages/sparo_acronyms}
\usepackage{Packages/sparo_math}
\usepackage{Packages/sparo_SIunits}
\usepackage{Packages/sparo_misc}
\usepackage{multirow}
\usepackage{indentfirst}
\usepackage{makecell}

\usepackage{soul,color}
\usepackage{verbatim} 
\usepackage{colortbl}
\usepackage[font=small,skip=4pt]{caption}
\usepackage{float} 

\usepackage{xcolor}
\usepackage[table]{xcolor}

\definecolor{gl}{HTML}{008000}

\title{\LARGE \bf
    $\sqrt{\text{VINS}}$-Based 360 Visual-Inertial SLAM Systems:\\
    Hilti--Trimble SLAM Challenge 2026 Entry
    \vspace{-0.7em}
}
\author{
    Jungwoo Lee$^{1}$, Hogyun Kim$^{1}$, Jiwon Choi$^{1}$, Jinyong Jeong$^{2}$, and Younggun Cho$^{1\dagger}$ \\
    {\small $^{1}$Dept. of Electrical and Computer Engineering, Inha University, Incheon, South Korea}\\
    {\small $^{2}$Riibotics, Siheung, South Korea} \\
    {\small {\tt [pihsdneirf, hg.kim, jiwon2]@inha.edu, jinyong.jeong@riibotics.io, yg.cho@inha.ac.kr}}
    \vspace{-3em}
}

\begin{document}
\maketitle
\input{PaperWriting/0_abstract.tex}
\input{PaperWriting/1_method.tex}
\input{PaperWriting/2_experiments.tex}

\small
\bibliographystyle{Packages/IEEEtranN} 
\balance
\bibliography{Packages/string-short, Packages/references}

\end{document}

%% file: PaperWriting/0_abstract.tex
\begin{strip}
    \centering
        {%
		\includegraphics[clip, width=0.98\textwidth]{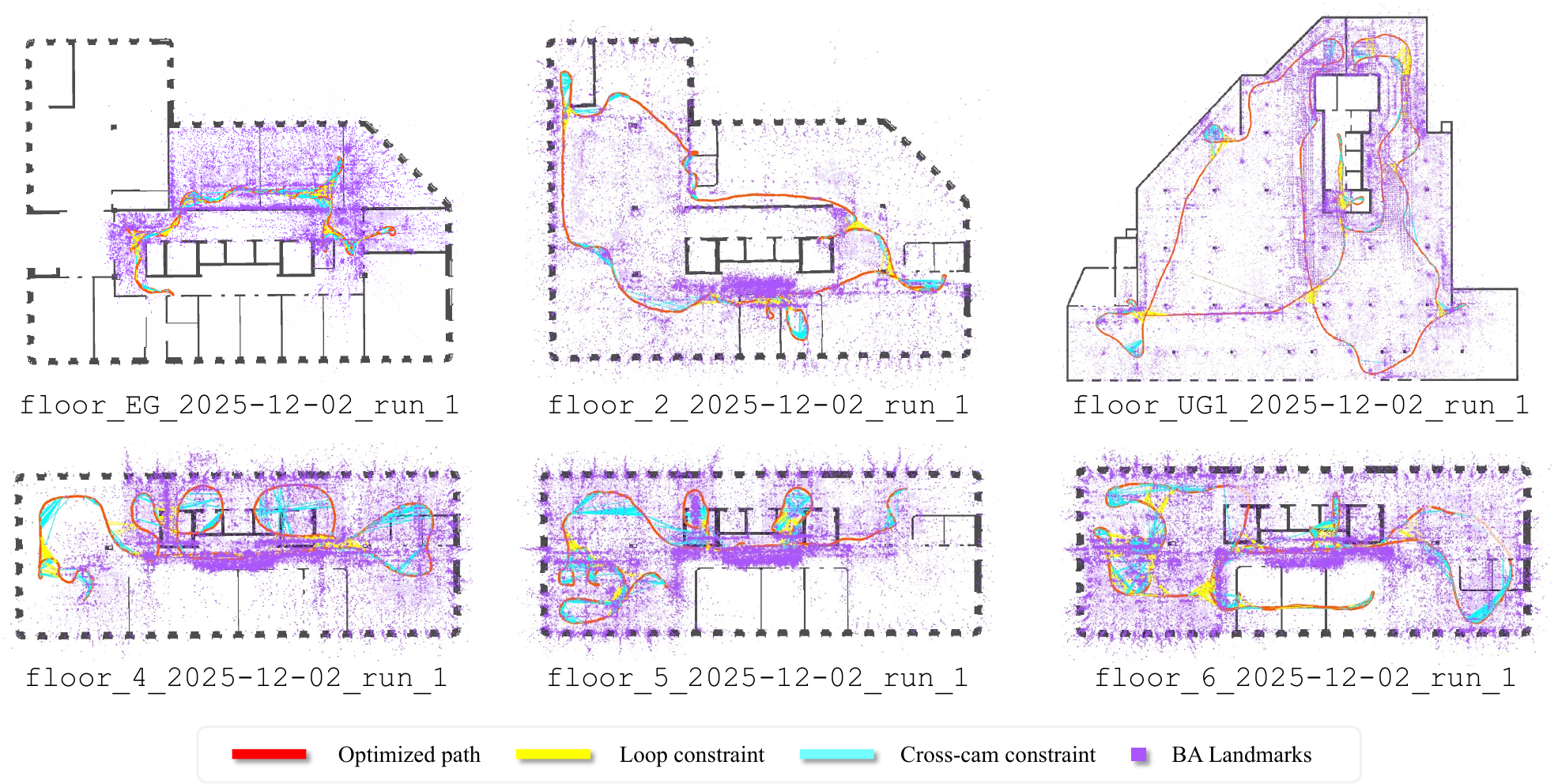}
	  }
    \captionof{figure}{Qualitative results on several Hilti-Trimble indoor sequences~\cite{slamchallenge2026}. 
                       Our simultaneous localization and mapping~(SLAM)~algorithm produces bundle adjustment~(BA)-optimized landmarks that align well with the floorplan map. 
                       Note that the alignment between the estimated map and the floorplan was performed manually for visualization.}    
\label{fig:pose_plots}
\vspace{-0.4cm}
\end{strip}

\begin{abstract}
This report describes our submission to the 2026 Hilti-Trimble SLAM Challenge. 
The system is a 360-camera visual-inertial pipeline that uses only the provided dual-fisheye image streams and IMU measurements. 
First, raw images are preprocessed with low-light enhancement~\cite{zdcepp} and static/dynamic masking~\cite{yolo26}, and then fed into a $\sqrt{\text{VINS}}$~\cite{srvins} frontend as the local odometry, with deep learning-based SuperPoint features~\cite{superpoint}, ELSED line segments~\cite{elsed}, and LightGlue matching~\cite{lightglue}. 
The resulting trajectory is refined using a GTSAM-based backend that performs tightly-coupled factor graph optimization with SALAD global descriptor~\cite{salad} and visual bundle adjustment~(BA)~\cite{smartfactor}. 
The backend performs local-to-global structureless BA, tightly regularized intrinsic and extrinsic refinement, and global scale adjustment supported by the known inter-camera baseline and cross-view constraints. 
The design targets long indoor sequences with dark regions, repetitive corridors, dynamic objects, and limited geometric parallax, while keeping the pipeline reproducible in the provided ROS2 workspace.
Fig.~\ref{fig:pose_plots} demonstrates that our proposed method aligns well with the floorplan across several indoor sequences.
\end{abstract}

%% file: PaperWriting/1_method.tex
\section{Algorithm Overview}
\label{sec:algorithm_overview}
\begin{figure*}[t]
	\centering
        {%
		\includegraphics[clip, width=\textwidth]{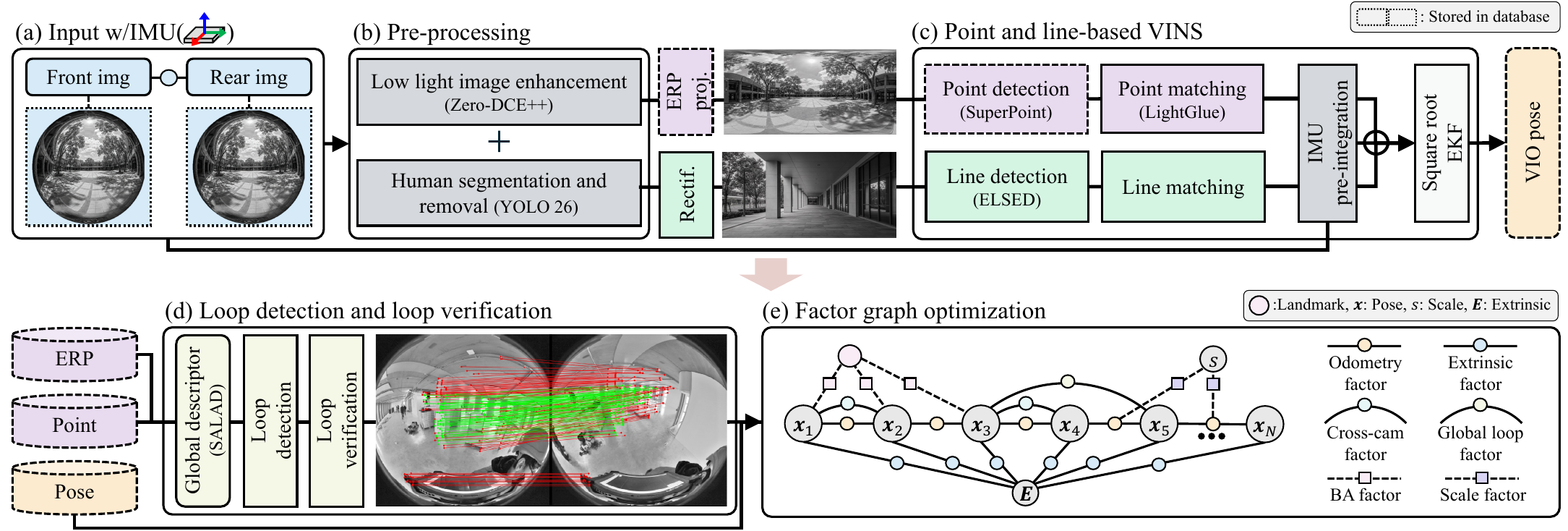}
	  }
    \vspace{-0.4cm}
    \caption{Overview of the proposed 360-camera visual-inertial SLAM pipeline. 
         Raw dual-fisheye frames are first conditioned by low-light enhancement~\cite{zdcepp} and dynamic masking~\cite{yolo26}, then tracked by a \(\sqrt{\mathrm{VINS}}\)-based frontend~\cite{srvins} with SuperPoint-LightGlue point features~\cite{superpoint,lightglue} and ELSED line segments~\cite{elsed}. 
         Detected loop closures~\cite{salad} and short-term cross-camera matches are geometrically verified and inserted as smart projection factors~\cite{smartfactor} into a GTSAM backend~\cite{isam2} that jointly optimizes poses, calibration, and a global scale variable.}
     \vspace{-0.5cm}
\label{fig:pipeline}
\end{figure*}

Our visual-inertial SLAM system is summarized in Fig.~\ref{fig:pipeline}. 
We model the 360 camera as two calibrated fisheye views under the extended and unified camera model~(EUCM) projection model and represent all poses in the \texttt{cam0} frame~(Sec.~\ref{sec:sensor_model}). 
First, raw fisheye frames are conditioned by static masking, low-light enhancement, and dynamic-object suppression before any feature extraction~(Sec.~\ref{sec:preprocessing}). 
Next, we pass the conditioned images to a \(\sqrt{\mathrm{VINS}}\)-based frontend~\cite{srvins} that estimates locally metric visual-inertial odometry using SuperPoint point features~\cite{superpoint}, LightGlue matching~\cite{lightglue}, and ELSED line segments~\cite{elsed}~(Sec.~\ref{sec:frontend}). 
We then detect and geometrically verify loop closure and short-term cross-camera constraints using SALAD global descriptor~\cite{salad} before inserting them into the factor graph~(Sec.~\ref{sec:loop}). 
Finally, a GTSAM-based backend~\cite{isam2} then jointly optimizes \texttt{cam0} poses, extrinsic and intrinsic parameters, and a global scale variable via EUCM-compatible structureless smart projection factors~\cite{smartfactor}~(Sec.~\ref{sec:backend}). 
Since loop closure and BA can revise earlier poses using later observations, the exported trajectory is non-causal.

\subsection{Sensor Model and Input Conditioning}
\label{sec:sensor_model}
First, we model the 360 camera as two calibrated fisheye views using the EUCM~\cite{eucm}. 
The optimized trajectory is represented in the \texttt{cam0} frame, where each graph node \(\mathbf{T}_k \in \mathrm{SE}(3)\) denotes the \texttt{cam0} pose at frame \(k\). 
The \texttt{cam0} observations are projected directly from \(\mathbf{T}_k\), while \texttt{cam1} observations are projected through a shared rigid transform \(\mathbf{E}_{01}=\mathbf{T}_{\mathrm{cam0}\rightarrow\mathrm{cam1}}\), allowing both fisheye views to constrain one \texttt{cam0} trajectory without duplicate per-frame camera nodes.

\subsection{Pre-processing}
\label{sec:preprocessing}
Next, before feature extraction, we apply three conditioning steps to each raw fisheye frame. 
Static masks remove invalid fisheye regions and fixed image artifacts. 
Low-light enhancement~\cite{zdcepp} then improves contrast in dark indoor regions. 
Finally, YOLO26 detector/segmentation outputs~\cite{yolo26} conservatively suppress features on potentially moving objects.

\subsection{Point- and line-based Visual-Inertial Odometry}
\label{sec:frontend}
We then build the frontend on \(\sqrt{\mathrm{VINS}}\)~\cite{srvins}. 
Point tracking uses SuperPoint~\cite{superpoint} and LightGlue~\cite{lightglue} on the conditioned fisheye images, with state-predicted matching to improve robustness during fast motion and repetitive indoor appearance. 
To complement point features in low-texture indoor regions, we add a line-feature branch that detects line segments with ELSED~\cite{elsed} and maintains their temporal associations via line matching. 
These line observations are incorporated through an MSCKF-style update inspired by PL-VIWO~\cite{plviwo}. 
Rather than enforcing a full Manhattan-world model, we adopt a vertical-world, or weak Manhattan-world, assumption: gravity provides a reliable vertical reference, while horizontal directions remain unconstrained.

\subsection{Loop Closure and Cross-View Constraints}
\label{sec:loop}
Following odometry estimation, loop closure uses global image descriptors~\cite{salad} to propose visually similar keyframes while rejecting immediate temporal neighbors, followed by SuperPoint-LightGlue geometric verification over valid \texttt{cam0}/\texttt{cam1} view pairings. 
We add accepted loop candidates as EUCM-compatible smart projection factors, keeping loop closure consistent with the keyframe BA formulation. 
Beyond long-range loop closure, we search a bounded temporal window for short-term cross-camera matches that connect nearby non-consecutive observations across the two fisheye views. 
These constraints are particularly useful during turns, corridor motion, and near-revisits, where structure observed in one fisheye view quickly reappears in the other. 
Because the verified matches span the calibrated inter-camera baseline, they also provide local metric evidence that supports the graph-wide scale correction.

\subsection{Factor Graph Optimization}
\label{sec:backend}
Finally, the backend optimizes a factor graph over \texttt{cam0} poses, camera intrinsic and extrinsic parameters, and a global scale variable. 
Consecutive frontend estimates provide relative pose factors, while all visual constraints are represented as EUCM-compatible structureless smart projection factors~\cite{smartfactor} instantiated for keyframe BA, geometrically verified loop closure, and short-term cross-camera tracks.
The backend objective is summarized as follows:
\begin{subequations}
\begin{align}
    \min_{\{\mathbf{T}_k\},\mathbf{E}_{01},\boldsymbol{\theta},s} \;
    &\sum_{(i,j)\in\mathcal{O}}
    \left\| r^{\mathrm{odom}}_{i,j}(\mathbf{T}_i,\mathbf{T}_j,s) \right\|^2_{\Sigma_o} \\
    &+ \sum_{m\in\mathcal{B}}
    \left\| r^{\mathrm{ba}}_{m}(\{\mathbf{T}\},\mathbf{E}_{01},\boldsymbol{\theta}) \right\|^2_{\Sigma_b} \\
    &+ \sum_{m\in\mathcal{L}}
    \left\| r^{\mathrm{loop}}_{m}(\{\mathbf{T}\},\mathbf{E}_{01},\boldsymbol{\theta}) \right\|^2_{\Sigma_l} \\
    &+ \sum_{m\in\mathcal{C}}
    \left\| r^{\mathrm{cross}}_{m}(\{\mathbf{T}\},\mathbf{E}_{01},\boldsymbol{\theta}) \right\|^2_{\Sigma_c} \\
    + \left\| r^{\mathrm{ext}}(\mathbf{E}_{01}) \right\|^2_{\Sigma_e}
    &+ \left\| r^{\mathrm{int}}(\boldsymbol{\theta}) \right\|^2_{\Sigma_{\theta}}
    + \left\| r^{\mathrm{scale}}(s) \right\|^2_{\Sigma_s},
\end{align}
\end{subequations}
where \(\mathcal{O}\) is the set of consecutive odometry edges, \(\mathcal{B}\) is the set of keyframe BA tracks, \(\mathcal{L}\) is the set of accepted loop tracks, and \(\mathcal{C}\) is the set of short-term cross-camera tracks. 
The residual \(r^{\mathrm{odom}}\) is a relative \(\mathrm{SE}(3)\) constraint whose translation is modulated by the shared scale variable \(s\), while \(r^{\mathrm{ba}}\), \(r^{\mathrm{loop}}\), and \(r^{\mathrm{cross}}\) are all EUCM-compatible smart projection residuals. 
We tightly regularize the shared extrinsic, intrinsic, and scale variables so that they absorb small residual calibration or metric-scale errors without accommodating arbitrary trajectory drift.

\begin{table*}[!t]
\centering
\caption{Transcription table matching the official evaluator report. 
        \xmark{} indicates that the sequence is excluded from evaluation because it is used as the ground truth trajectory.
        Note that \textcolor{red!8}{\rule[-0.1em]{1.8em}{0.9em}} indicates hidden plots.}
\label{tab:evaluator_report_transcription}
\tiny
\setlength{\tabcolsep}{6pt}
\renewcommand{\arraystretch}{0.9}
\resizebox{\textwidth}{!}{%
\begin{tabular}{lccccccc}
\toprule
\midrule
Sequence & RMSE [m] & Mean [m] & Max [m] & Min [m] & Coverage [\%] & Poses & Exp. Score \\
\midrule
\texttt{floor\_1\_2025-05-05\_run\_1}                    & \xmark & \xmark & \xmark & \xmark & \xmark   & \xmark & \xmark \\
\rowcolor{red!8} \texttt{floor\_1\_2025-07-07\_run\_1}   & 0.1465 & 0.1264 & 0.4078 & 0.0066 & 100.0000 & 3959   & 94.4015 \\
\texttt{floor\_1\_2025-12-02\_run\_1}                    & 0.1060 & 0.0969 & 0.2190 & 0.0100 & 100.0000 & 8172   & 95.6537 \\
\texttt{floor\_2\_2025-05-05\_run\_1}                    & \xmark & \xmark & \xmark & \xmark & \xmark   & \xmark & \xmark \\
\texttt{floor\_2\_2025-10-28\_run\_1}                    & \xmark & \xmark & \xmark & \xmark & \xmark   & \xmark & \xmark \\
\texttt{floor\_2\_2025-10-28\_run\_2}                    & \xmark & \xmark & \xmark & \xmark & \xmark   & \xmark & \xmark \\
\rowcolor{red!8} \texttt{floor\_2\_2025-12-02\_run\_1}   & 0.1354 & 0.1267 & 0.5778 & 0.0109 & 99.9304  & 4306   & 94.3535 \\
\texttt{floor\_2\_2025-12-03\_run\_1}                    & 0.1883 & 0.1534 & 0.6852 & 0.0097 & 99.9321  & 4413   & 93.2931 \\
\texttt{floor\_3\_2025-05-19\_run\_1}                    & 0.0907 & 0.0807 & 0.1867 & 0.0063 & 100.0000 & 3383   & 96.3713 \\
\texttt{floor\_3\_2025-12-02\_run\_1}                    & 0.0866 & 0.0778 & 0.2001 & 0.0042 & 99.9229  & 3890   & 96.4937 \\
\texttt{floor\_4\_2025-05-19\_run\_1}                    & 0.0783 & 0.0732 & 0.1377 & 0.0118 & 100.0000 & 2616   & 96.6919 \\
\texttt{floor\_4\_2025-12-02\_run\_1}                    & 0.0918 & 0.0816 & 0.2329 & 0.0042 & 97.0816  & 5522   & 96.3304 \\
\texttt{floor\_5\_2025-12-02\_run\_1}                    & 0.0837 & 0.0731 & 0.2366 & 0.0065 & 99.9366  & 4728   & 96.7088 \\
\texttt{floor\_6\_2025-06-18\_run\_1}                    & 0.0588 & 0.0474 & 0.3076 & 0.0097 & 100.0000 & 2029   & 97.8519 \\
\rowcolor{red!8} \texttt{floor\_6\_2025-07-07\_run\_1}   & 0.0693 & 0.0629 & 0.1266 & 0.0096 & 100.0000 & 2104   & 97.1545 \\
\texttt{floor\_6\_2025-12-02\_run\_1}                    & 0.0883 & 0.0757 & 0.2253 & 0.0052 & 100.0000 & 5041   & 96.5962 \\
\texttt{floor\_6\_2025-12-02\_run\_2}                    & 0.0882 & 0.0741 & 0.2662 & 0.0026 & 99.9180  & 3654   & 96.6706 \\
\texttt{floor\_7\_2025-12-02\_run\_1}                    & 0.0889 & 0.0647 & 0.3414 & 0.0015 & 99.9138  & 3476   & 97.1012 \\
\texttt{floor\_7\_2025-12-02\_run\_2}                    & 0.0749 & 0.0687 & 0.1619 & 0.0027 & 100.0000 & 4539   & 96.8927 \\
\rowcolor{red!8} \texttt{floor\_7\_2025-12-03\_run\_1}   & 0.1134 & 0.0992 & 0.3443 & 0.0078 & 100.0000 & 5877   & 95.5658 \\
\rowcolor{red!8} \texttt{floor\_EG\_2025-10-16\_run\_1}  & 0.0987 & 0.0854 & 0.2563 & 0.0012 & 100.0000 & 7170   & 96.1669 \\
\texttt{floor\_EG\_2025-12-02\_run\_1}                   & 0.0770 & 0.0611 & 0.3792 & 0.0038 & 99.9191  & 3707   & 97.2465 \\
\texttt{floor\_EG\_2025-12-02\_run\_2}                   & 0.0667 & 0.0601 & 0.1976 & 0.0038 & 99.8538  & 4781   & 97.2796 \\
\texttt{floor\_UG1\_2025-05-19\_run\_1}                  & 0.1318 & 0.1159 & 0.2886 & 0.0197 & 100.0000 & 6010   & 94.8434 \\
\texttt{floor\_UG1\_2025-06-18\_run\_1}                  & 0.1897 & 0.1653 & 0.4832 & 0.0153 & 100.0000 & 4889   & 92.7524 \\
\texttt{floor\_UG1\_2025-10-16\_run\_1}                  & \xmark & \xmark & \xmark & \xmark & \xmark   & \xmark & \xmark \\
\texttt{floor\_UG1\_2025-12-02\_run\_1}                  & 0.1616 & 0.1457 & 0.3578 & 0.0242 & 99.9603  & 7548   & 93.5574 \\
\texttt{floor\_UG1\_2025-12-02\_run\_2}                  & 0.1643 & 0.1396 & 0.4202 & 0.0041 & 99.9846  & 6501   & 93.8487 \\
\texttt{floor\_UG1\_2025-12-03\_run\_1}                  & 0.1093 & 0.1024 & 0.2307 & 0.0101 & 99.8490  & 3967   & 95.4095 \\
\rowcolor{red!8} \texttt{floor\_UG2\_2025-12-02\_run\_1} & 0.1367 & 0.1200 & 0.3073 & 0.0137 & 100.0000 & 6607   & 94.6656 \\
\midrule
\textbf{Mean Results} & \textbf{0.1090} & \textbf{0.0951} & \textbf{0.3031} & \textbf{0.0082} & \textbf{99.8481} & & \\
\midrule
\textbf{Final Score} & & & & & & & \textbf{2393.9009} \\
\midrule
\bottomrule
\end{tabular}%
}
\end{table*}

%% file: PaperWriting/2_experiments.tex
\section{Experiments}
\label{sec:experiments}

Table~\ref{tab:evaluator_report_transcription} shows that our method achieves a mean position error below 0.1\,m across the evaluated sequences.